\documentclass[conference]{IEEEtran}

\usepackage{cite}
\usepackage{amsmath,amssymb,amsfonts}
\usepackage{graphicx}
\usepackage{textcomp}
\usepackage{xcolor}
\usepackage{booktabs}

\graphicspath{{figures/final/}{figures/archive/}}

\def\BibTeX{{\rm B\kern-.05em{\sc i\kern-.025em b}\kern-.08em
    T\kern-.1667em\lower.7ex\hbox{E}\kern-.125emX}}

\newif\ifanonymous
\anonymousfalse

\begin{document}
\bstctlcite{IEEEtranBSTCTL:etal}

\title{MIFLOW: Dynamic Recent-Behavior Priors for Reference-Calibrated Online Flow-Policy Reinforcement Learning}

\ifanonymous
\author{\IEEEauthorblockN{Anonymous Authors}}
\else
\author{
\IEEEauthorblockN{
Chenhao Wang\textsuperscript{1},
Haonan Ma\textsuperscript{1},
Wenjie Zhang\textsuperscript{2},
Yifan Wang\textsuperscript{2},\\
Mingbin Gao\textsuperscript{1,*},
Yuan-Ye Zhang\textsuperscript{1,*},
and Wenjing Hong\textsuperscript{1,*}}
\IEEEauthorblockA{\textsuperscript{1}Institute of Artificial Intelligence, Xiamen University, Xiamen, China}
\IEEEauthorblockA{\textsuperscript{2}Xiamen University, Xiamen, China}
\IEEEauthorblockA{chenhao@stu.xmu.edu.cn; mhn\_xmu@163.com; zhangwenjie23@xmu.edu.cn; yifan@xmu.edu.cn\\
mbgao@xmu.edu.cn; zhangyuanye@xmu.edu.cn; whong@xmu.edu.cn}
\IEEEauthorblockA{\textsuperscript{*}Corresponding authors: Mingbin Gao, Yuan-Ye Zhang, and Wenjing Hong}
}
\fi

\AddToHookNext{shipout/foreground}{%
  \put(\dimexpr1in+\hoffset+\oddsidemargin\relax,%
       \dimexpr-\paperheight+0.67in\relax){%
    \makebox(0,0)[lt]{%
      \parbox[t]{\textwidth}{\footnotesize\raggedright
        \textcopyright~2026 IEEE. Personal use of this material is permitted.
        Permission from IEEE must be obtained for all other uses, in any current
        or future media, including reprinting/republishing this material for
        advertising or promotional purposes, creating new collective works, for
        resale or redistribution to servers or lists, or reuse of any
        copyrighted component of this work in other works.%
      }%
    }%
  }%
}
\maketitle

\begin{abstract}
Flow-based policies provide expressive action distributions for continuous-control reinforcement learning, and they offer a flexible alternative to unimodal Gaussian actors. Online actor-critic learning updates the policy and critic from finite, evolving interaction data. However, off-policy value estimates can become unreliable in poorly covered action regions, and this extrapolation risk can be amplified by flow-based policies whose expressive generators can rapidly shift probability mass toward critic-favored actions with limited direct interaction data. Therefore, we propose MIFLOW, a reference-calibrated online flow-policy learning method that constructs a behavior-induced reference from recent interactions to guide value-driven policy improvement. Motivated by canalization as a qualitative analogy for path-dependent stabilization, MIFLOW learns a recency-adaptive flow-matching prior that summarizes the empirical action distribution induced by recent interactions. Rather than imposing a hard behavior constraint, this prior defines a moving density reference that shapes policy improvement: actions assigned low density by the recent-behavior prior are not forbidden, but incur an additional cost during value-guided policy improvement. The learned prior induces an approximate recent-action log-density, which is used in a dimension-normalized reference correction integrated into both actor updates and critic targets. Experiments on six MuJoCo continuous-control tasks using five random seeds report competitive mean return curves and higher late-training mean returns than the evaluated baselines on several tasks. In a task-specific diagnostic case study, MIFLOW reports lower fractions of policy actions, including critic-favored high-Q actions, in regions assigned low density by the learned reference. These diagnostics measure alignment with the learned reference, not true support or critic extrapolation error. Prior-family and prior-buffer comparisons produced task-dependent differences in return and prior-defined diagnostics. These results support dynamic recent-behavior priors as a practical calibration mechanism in the evaluated MuJoCo settings, but do not constitute a direct measurement or guarantee of reduced critic extrapolation.
\end{abstract}


\section{Introduction}

Entropy-regularized actor-critic methods such as soft actor-critic (SAC) are standard tools for continuous-control reinforcement learning, typically parameterizing the actor as a unimodal Gaussian for tractable stochastic optimization \cite{haarnoja2018soft,haarnoja2019softactorcriticalgorithmsapplications}. Recent diffusion- and flow-based policies provide richer classes that can capture complex or multimodal action distributions \cite{wang2023diffusion,psenka2024qscore,celik2025dime,park2025flowq}. In online actor-critic learning, however, policy improvement is driven by critic estimates from finite, nonstationary interaction data. Value estimates can be unreliable in poorly covered action regions, and expressive flow generators can rapidly reallocate probability mass toward critic-favored actions weakly represented in recent interaction data. The resulting challenge is reference-calibrated policy improvement that retains policy expressivity while calibrating value-guided updates against recent empirical action coverage.

Out-of-support value extrapolation is a well-known failure mode in offline and fixed-batch off-policy reinforcement learning \cite{levine2020offline,yu2020mopo,yu2021combo}. Online interaction can revise erroneous estimates, but function-approximation and overestimation errors can still bias policy improvement when the actor changes faster than new data provide corrective coverage \cite{fujimoto2018td3}. Behavior-constrained, safe-improvement, and implicit policy-improvement methods limit this risk by restricting policy changes or reducing deviation from dataset behavior \cite{laroche2019safe,fujimoto2021minimalist,kostrikov2022offline}. Directly transferring a fixed-support view is insufficient for online flow policies because their policy-induced data distribution evolves throughout training. They instead require a recency-adaptive reference that tracks recent behavior while remaining soft enough to calibrate, rather than constrain, value-guided policy improvement.

Recent online flow-policy methods improve flow-policy optimization, path-density estimation, and rollout stability \cite{lv2025flowrl,zhang2025sac}. Although they establish that iterative flow policies can be trained within online actor-critic frameworks, trainability alone does not calibrate critic-guided updates against recent empirical action coverage. The central problem is therefore how to avoid amplifying value extrapolation in poorly covered action regions while preserving the expressive policy class.

To address this problem, we propose MIFLOW, a reference-calibrated online flow-policy method that uses a behavior-induced reference from recent interactions. Canalization provides a qualitative analogy for path-dependent adaptation, and MIFLOW learns a recency-adaptive flow-matching prior of the recent action distribution \cite{waddington1942canalization,gibson2000canalization}. A dimension-normalized, mutual-information-motivated reference correction applies an additional cost to critic-favored actions with low prior density \cite{leibfried2020mutual}. Such actions are not forbidden. The prior is not a hard behavior constraint, an exact current-policy marginal, or a formal robustness guarantee; it is a soft density reference for value-guided policy improvement. Experiments on six MuJoCo continuous-control tasks using five random seeds report task-dependent competitive or higher mean returns relative to the evaluated baselines. A task-specific diagnostic case study reports lower fractions of policy actions, including critic-favored high-Q actions, in regions assigned low density by the learned reference. These diagnostics measure policy-prior alignment rather than true support or critic extrapolation error.

\section{Related Work}

\noindent\textbf{Generative and flow policies in reinforcement learning.}
Building on diffusion-model foundations \cite{ho2020ddpm}, generative policies provide expressive alternatives to Gaussian actors. Diffusion-QL combines conditional diffusion behavior modeling with value maximization \cite{wang2023diffusion}. Q-score matching connects diffusion-policy scores to Q-function action gradients \cite{psenka2024qscore}, whereas DIME integrates diffusion policies into maximum-entropy reinforcement learning \cite{celik2025dime}. DACER develops an online diffusion actor-critic with entropy regulation \cite{wang2024dacer}. Flow Q-Learning uses a flow-matching policy and an expressive one-step policy in offline and offline-to-online settings \cite{park2025flowq}. Among recent online flow-based methods, FlowRL studies Wasserstein-regularized flow-policy optimization \cite{lv2025flowrl}. ReinFlow fine-tunes pretrained flow-matching policies using likelihood-aware stochasticity \cite{zhang2025reinflow}, and SACFlow constructs a tractable path-density actor-critic objective \cite{zhang2025sac}. MIFLOW complements these methods by learning a recent-behavior prior that calibrates critic-guided path-density updates.

\vspace{0.35em}
\noindent\textbf{Reference regularization, behavior support, and path-dependent stabilization.}
Reference distributions and support constraints provide complementary ways to regulate policy improvement. TRPO uses a trust-region objective \cite{schulman2015trpo}, whereas PPO uses a clipped probability-ratio surrogate \cite{schulman2017ppo}. MPO and V-MPO formulate value-weighted improvement through relative-entropy-style constraints \cite{abdolmaleki2018mpo,song2020vmpo}. Mutual-information-regularized actor-critic learning introduces a state-independent marginal action reference and a policy-reference log-ratio \cite{leibfried2020mutual}. In offline reinforcement learning, BCQ restricts action selection using a learned behavior model \cite{fujimoto2019bcq}. BEAR limits policy deviation from dataset support \cite{kumar2019bear}, whereas CQL reduces value optimism through conservative value learning \cite{kumar2020cql}. Online flow-policy learning instead involves a policy-induced data distribution that changes with the critic and actor. MIFLOW adapts reference regularization to this setting using a recency-adaptive prior learned from recent interactions. Canalization provides only a qualitative analogy for this path-dependent adaptation \cite{waddington1942canalization,gibson2000canalization}; the algorithm is defined by the learned reference and actor-critic objectives.

\section{Preliminaries}

\subsection{Entropy-Regularized Off-Policy Actor-Critic}

We consider an infinite-horizon Markov decision process defined by continuous
state and action spaces \(\mathcal{S}\) and \(\mathcal{A}\), transition kernel
\(P\), and reward function \(r\). Returns are aggregated using a discount
factor \(\gamma\in[0,1)\). In off-policy actor-critic learning,
transitions \((s,a,r,s',d)\) are stored in a replay buffer \(\mathcal{D}\), where
\(d\in\{0,1\}\) denotes the bootstrap-termination mask: \(1-d\) retains the
next-state target only when bootstrapping is allowed.

SAC augments value-guided policy improvement with an entropy
regularizer \cite{haarnoja2018soft,haarnoja2019softactorcriticalgorithmsapplications}.
With clipped double critics \(Q_{\psi_i}\), target critics \(Q_{\bar{\psi}_i}\),
and entropy temperature \(\alpha\), the soft critic target and actor objective
can be written separately. The critic target is
\begin{equation}
\begin{aligned}
y_{\mathrm{SAC}}
&=
r+\gamma(1-d)
\left[
\min_i Q_{\bar{\psi}_i}(s',a') \right.\\
&\quad\left.
-
\alpha\log\pi_\theta(a'|s')
\right],
\end{aligned}
\label{eq:sac_background}
\end{equation}
where \(a'\sim\pi_\theta(\cdot|s')\). The actor objective is
\begin{equation}
\begin{aligned}
\mathcal{L}_{\mathrm{actor}}
&=
\mathbb{E}_{\substack{s\sim\mathcal{D}\\a\sim\pi_\theta(\cdot|s)}}
\left[
\alpha\log\pi_\theta(a|s) \right.\\
&\quad\left.
-
\min_i Q_{\psi_i}(s,a)
\right],
\end{aligned}
\label{eq:sac_actor_background}
\end{equation}
The key quantity for entropy-regularized
actor-critic learning is therefore a computable policy log-density. For
multi-step flow policies, this term is not directly available as a simple
Gaussian log-probability, motivating the path-density formulation reviewed
next.

\subsection{Flow Policies and SACFlow Path Density}

Flow-based policies generate continuous actions by transforming a simple base
distribution through a state-conditioned velocity field. Let
\begin{equation}
0=t_0<t_1<\cdots<t_K=1
\end{equation}
denote discretized flow time points. Starting from a base action variable
\begin{equation}
A_{t_0}\sim\mathcal{N}(0,I),
\end{equation}
a flow policy generates intermediate action variables according to
\begin{equation}
A_{t_{i+1}}
=
A_{t_i}
+
\Delta t_i v_\theta(t_i,A_{t_i},s),
\qquad
i=0,\ldots,K-1,
\label{eq:flow_rollout}
\end{equation}
where \(v_\theta\) is the velocity field and
\(\Delta t_i=t_{i+1}-t_i\). The final bounded environment action is
\begin{equation}
a=\tanh(A_{t_K}),
\end{equation}
and the full flow path is denoted by
\begin{equation}
A=(A_{t_0},A_{t_1},\ldots,A_{t_K}).
\end{equation}

Directly applying entropy-regularized actor-critic learning to this iterative
generation process is difficult because the marginal likelihood of the final
action can be intractable. SACFlow addresses this issue by using a
noise-augmented rollout that induces a tractable joint density over the flow
path \cite{zhang2025sac}. We write
\begin{equation}
\begin{aligned}
A\sim\pi_\theta^c(\cdot|s),
&\qquad
p_\theta^c(A|s)
\end{aligned}
\end{equation}
for the corresponding path distribution and its tractable density. The term
\(\log p_\theta^c(A|s)\) can then be used as a path-level surrogate for the
policy log-density in SAC-style optimization.

For example, the SACFlow actor objective can be written as
\begin{equation}
\begin{aligned}
\mathcal{L}^{\mathrm{SACFlow}}_{\mathrm{actor}}
&=
\mathbb{E}_{\substack{s\sim\mathcal{D}\\A\sim\pi_\theta^c(\cdot|s)}}
\left[
\alpha\log p_\theta^c(A|s) \right.\\
&\quad\left.
-
\min_i Q_{\psi_i}(s,a)
\right],
\end{aligned}
\label{eq:sacflow_actor_background}
\end{equation}
where \(a=\tanh(A_{t_K})\). This path-density formulation provides the
density term needed to train iterative flow policies in an entropy-regularized
actor-critic framework.

\subsection{Reference-Regularized Policy Improvement}

Reference-based regularization provides a general way to move beyond
entropy-only policy regularization. In mutual-information regularized
actor-critic learning, a state-conditioned policy is compared with a
state-independent marginal action distribution through a policy-reference
log-ratio \cite{leibfried2020mutual}:
\begin{equation}
\log\pi(a|s)-\log\pi_{\mathrm{marginal}}(a).
\label{eq:mi_log_ratio_background}
\end{equation}
Here, \(\pi_{\mathrm{marginal}}(a)\) denotes the marginal action distribution
induced by the policy and a state distribution. This perspective suggests that
policy improvement can be regularized relative to an action reference, rather
than only through conditional policy entropy.

For a fixed reference distribution \(\rho(a)\), the expected unweighted
log-ratio corresponds to a KL-to-reference regularizer:
\begin{equation}
\begin{aligned}
\mathbb{E}_{s}
D_{\mathrm{KL}}
\left(
\pi(\cdot|s)
\middle\|
\rho(\cdot)
\right)
=
\mathbb{E}_{s,a\sim\pi}
\left[
\log\pi(a|s)-\log\rho(a)
\right].
\end{aligned}
\label{eq:kl_reference_background}
\end{equation}
A corresponding reference-regularized actor objective has the generic form
\begin{equation}
\mathbb{E}_{s,a}
\left[
\alpha
\left(
\log\pi(a|s)-\log\rho(a)
\right)
-
Q(s,a)
\right].
\label{eq:reference_actor_background}
\end{equation}

We use this MI/reference-regularization perspective only as motivation. The
method developed next adapts this idea to online flow-policy learning, where
the conditional policy term is represented by a tractable flow-path density
and the reference is learned from recent behavior rather than assumed to be an
exact current-policy marginal.

\section{Method}

\subsection{Overview: Dynamic Reference Calibration for Online Flow Policies}

\begin{figure}[t]
\centering
\includegraphics[width=\columnwidth]{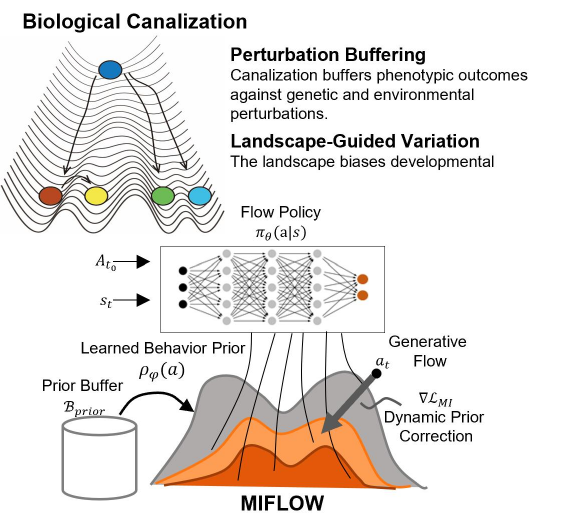}
\caption{
Overview of MIFLOW.
Recent executed actions are stored in a sliding prior buffer and used to train a recency-adaptive flow-matching behavior prior.
The prior provides an approximate recent-action log-density that acts as a dynamic reference in path-density actor-critic updates.
Low-prior-density actions remain selectable but receive an additional cost during critic-guided policy improvement.
The upper panel provides a qualitative analogy for path-dependent adaptation; the algorithm itself is defined by the learned prior and reference-calibrated objectives.
}
\label{fig:miflow_overview}
\end{figure}

MIFLOW is designed to calibrate critic-guided updates using a learned reference derived from recent interactions. Figure~\ref{fig:miflow_overview} summarizes its two coupled components: learning a recent-behavior reference and using that reference in path-density actor-critic updates. For a state $s$, the flow actor samples a generation path $A=(A_{t_0},\ldots,A_{t_K})\sim\pi_\theta^c(\cdot|s)$, produces the bounded action $a=\tanh(A_{t_K})$, and assigns a tractable path density $p_\theta^c(A|s)$. MIFLOW couples this path density with a recency-adaptive behavior prior $\rho_\phi(a)$ learned from recently executed actions.

The resulting dynamic-reference regularizer is
\begin{equation}
\label{eq:miflow_regularizer}
\begin{aligned}
\ell_{\mathrm{MIFLOW}}(s,A,a)
&=
\log p_\theta^c(A|s)
-\lambda_\rho \log \rho_\phi(a),\\
\lambda_\rho
&=\frac{1}{d_a},
\end{aligned}
\end{equation}
where $d_a$ is the action dimension. The path-density term preserves entropy-regularized optimization for the flow actor, while the prior-density term introduces a learned reference for recent action density. When $\rho_\phi(a)$ is low, the correction increases the effective cost of selecting $a$. Such actions remain selectable but require a larger value advantage to be preferred.

The action log-density contains a sum of contributions over $d_a$ action dimensions, so its magnitude generally increases with the action dimension. We therefore set $\lambda_\rho=1/d_a$ to express the prior correction on an approximately per-dimension scale. This normalization reduces systematic scale differences across environments with different action dimensions. It is a fixed heuristic rather than a theoretically optimal or empirically tuned coefficient, and we do not claim robustness to alternative values.

Canalization provides a qualitative analogy for path-dependent adaptation rather than a theoretical basis for MIFLOW. Technically, the method has two coupled steps: estimating recent action density and using that estimate to calibrate critic-guided actor-critic updates.

\subsection{Recency-Adaptive Flow-Matching Behavior Prior}

The dynamic-reference correction in MIFLOW requires a density reference that reflects the current stage of online interaction. To construct this reference, MIFLOW maintains a sliding prior buffer $\mathcal{B}_{\mathrm{prior},t}$ that stores recently executed actions. Unlike the replay buffer $\mathcal{D}$, which is used for off-policy value learning, $\mathcal{B}_{\mathrm{prior},t}$ is used only to estimate the recent marginal action density. This separation ensures that the reference summarizes what the agent has recently executed, rather than what the critic currently assigns high value.

MIFLOW learns a state-independent behavior prior
\begin{equation}
\rho_\phi(a),
\qquad
a\sim \mathcal{B}_{\mathrm{prior},t}.
\end{equation}
The prior models the marginal action density induced by recent behavior. Its parameters are updated from recent actions only, without Q-weighted targets or actor-gradient feedback. It therefore provides a recent-action density estimate that is operationally separated from the critic objective.

We parameterize $\rho_\phi(a)$ as a flow-matching density model over actions. For bounded actions, each action is first mapped to the pre-squashed space,
\begin{equation}
x_1=\operatorname{arctanh}(a).
\end{equation}
Given $x_0\sim\mathcal{N}(0,I)$ and flow time $\tau\sim\mathrm{Unif}([0,1])$, we construct the interpolation
\begin{equation}
x_\tau=(1-\tau)x_0+\tau x_1
\end{equation}
with target velocity
\begin{equation}
u_\tau=x_1-x_0.
\end{equation}
The prior velocity field $v_\phi$ is trained with the flow-matching objective
\begin{equation}
\label{eq:prior_loss}
\begin{aligned}
\mathcal{L}_{\mathrm{prior}}(\phi)
&=
\mathbb{E}_{\substack{a\sim\mathcal{B}_{\mathrm{prior},t}\\x_0,\,\tau}}
\left[
\left\|
v_\phi(\tau,x_\tau)-u_\tau
\right\|_2^2
\right].
\end{aligned}
\end{equation}

\vspace{0.25em}
\noindent\textit{Prior-density evaluation.}
The flow-matching loss in Eq.~\eqref{eq:prior_loss} trains the behavior-prior velocity field, but it does not by itself directly output an action log-density. For a bounded environment action $a$, we first normalize by the environment action bias $\mu_j$ and scale $\sigma_j$:
\begin{equation}
\hat a_j =
\frac{a_j-\mu_j}{\sigma_j+\varepsilon}.
\end{equation}
For numerical stability, normalized actions are clipped near the $[-1,1]$ boundary before applying the inverse tanh:
\begin{equation}
\tilde a_j =
\operatorname{clip}(\hat a_j,-1+\varepsilon,1-\varepsilon),
\qquad
x_1=\operatorname{arctanh}(\tilde a).
\end{equation}
In likelihood evaluation we use $\varepsilon=10^{-6}$. Let $p_\phi^X$ denote the density in the pre-squashed space. The learned prior velocity field defines
\begin{equation}
\frac{d x_t}{dt}=v_\phi(t,x_t),
\end{equation}
and the instantaneous change-of-variables formula gives
\begin{equation}
\frac{d\log p_t(x_t)}{dt}
=
-
\operatorname{Tr}
\left(
\frac{\partial v_\phi(t,x_t)}{\partial x_t}
\right).
\end{equation}
Thus,
\begin{equation}
\begin{aligned}
\log p_\phi^X(x_1)
&=
\log p_0(x_0)\\
&\quad
-
\int_0^1
\operatorname{Tr}
\left(
\frac{\partial v_\phi(t,x_t)}{\partial x_t}
\right)dt .
\end{aligned}
\end{equation}
In implementation, the backward integration is approximated with Euler steps, and the divergence term is estimated with a Hutchinson trace estimator,
\begin{equation}
\widehat{\operatorname{div}}\,v_\phi(t,x)
=
\eta^\top
\frac{\partial v_\phi(t,x)}{\partial x}
\eta,
\qquad
\eta\sim\mathcal{N}(0,I).
\end{equation}
The bounded-action log-density is then obtained by the tanh and action-scale change of variables:
\begin{equation}
\begin{aligned}
\log\rho_\phi(a)
&=
\log p_\phi^X(x_1)\\
&\quad
-
\sum_{j=1}^{d_a}
\left[
\log(\sigma_j+\varepsilon)
+
\log(1-\tilde a_j^2+\varepsilon)
\right].
\end{aligned}
\end{equation}
After training, the learned prior provides the log-density term $\log\rho_\phi(a)$ used in the dynamic-reference regularizer. Because $\mathcal{B}_{\mathrm{prior},t}$ is a sliding buffer, this density reference evolves with recent interaction data instead of remaining tied to a fixed dataset. The learned flow-matching prior can represent more flexible recent-action distributions than a Gaussian prior, while the Gaussian-prior variant is used only as an ablation baseline. We do not treat $\rho_\phi(a)$ as the exact current-policy marginal; it is an implementable recency-adaptive behavior reference for calibrating critic-guided updates. The resulting $\log\rho_\phi(a)$ is an approximate likelihood estimate under the learned behavior-prior flow rather than an exact likelihood or a direct output of the flow-matching loss.

\subsection{Reference-Calibrated Path-Density Actor-Critic}

Given the dynamic-reference regularizer in Eq.~\eqref{eq:miflow_regularizer}, MIFLOW calibrates both policy improvement and bootstrapped policy evaluation against the recent-behavior reference. During actor-critic updates, the prior parameters $\phi$ are held fixed and are updated only by the behavior-prior loss in Eq.~\eqref{eq:prior_loss}. Thus, $\rho_\phi(a)$ acts as a critic-decoupled behavior reference rather than a Q-weighted target.

The actor is optimized by minimizing
\begin{equation}
\label{eq:miflow_actor}
\begin{aligned}
\mathcal{L}^{\mathrm{MIFLOW}}_{\mathrm{actor}}
&=
\mathbb{E}_{\substack{s\sim\mathcal{D}\\A\sim\pi_\theta^c(\cdot|s)}}
\left[
\alpha\ell_{\mathrm{MIFLOW}}(s,A,a)
\right.\\
&\quad\left.
-\min_i Q_{\psi_i}(s,a)
\right],
\end{aligned}
\end{equation}
where $a=\tanh(A_{t_K})$. Compared with entropy-only path-density optimization, Eq.~\eqref{eq:miflow_actor} adds a recent-reference correction through $-\lambda_\rho\log\rho_\phi(a)$. When the critic assigns high value to an action with low prior density, the actor can still select that action, but its value advantage must overcome the additional cost. The present experiments characterize this effect relative to the learned prior rather than directly measuring critic extrapolation error.

For critic learning, given a transition $(s,a,r,s',d)\sim\mathcal{D}$, we sample a next flow path $A'\sim\pi_\theta^c(\cdot|s')$, set $a'=\tanh(A'_{t_K})$, and use the reference-calibrated target
\begin{equation}
\label{eq:miflow_target}
\begin{aligned}
y
&=
r+\gamma(1-d)
\Big[
\min_i Q_{\bar{\psi}_i}(s',a')\\
&\quad
-\alpha\ell_{\mathrm{MIFLOW}}(s',A',a')
\Big].
\end{aligned}
\end{equation}
The critic loss is
\begin{equation}
\label{eq:miflow_critic}
\begin{aligned}
\mathcal{L}_{\mathrm{critic}}
&=
\mathbb{E}_{(s,a,r,s',d)\sim\mathcal{D}}
\left[
\sum_i
\left(
Q_{\psi_i}(s,a)-y
\right)^2
\right].
\end{aligned}
\end{equation}
Using the same dynamic-reference term in the target makes policy evaluation consistent with the reference-calibrated actor update: low-prior-density next actions receive a lower entropy-regularized backup unless their value estimate is sufficiently large.

The temperature update follows the path-density entropy scale and does not include the prior correction:
\begin{equation}
\label{eq:alpha_loss}
\begin{aligned}
\mathcal{L}_{\alpha}
&=
\mathbb{E}_{s,A}
\left[
\alpha
\left(
-\log p_\theta^c(A|s)-\bar{\mathcal{H}}
\right)
\right],
\end{aligned}
\end{equation}
where $\bar{\mathcal{H}}$ is the target entropy. This keeps entropy calibration tied to the flow-path density, while the learned prior remains a separate reference-calibration term for critic-guided updates.

\subsection{Training Procedure and Dynamic-Reference Interpretation}

MIFLOW alternates online interaction, behavior-prior estimation, and reference-calibrated actor-critic learning. During environment interaction, each transition is stored in the replay buffer $\mathcal{D}$, and the executed action is additionally inserted into the sliding prior buffer $\mathcal{B}_{\mathrm{prior},t}$. The prior buffer is used only to update $\rho_\phi(a)$ with Eq.~\eqref{eq:prior_loss}, while the replay buffer is used for critic and actor updates. During actor-critic optimization, $\rho_\phi(a)$ is treated as a fixed recent-behavior reference; gradients from the Q-functions or the actor objective are not used to train the prior.

This decoupled training scheme separates three roles. The critic estimates value from replayed transitions, the flow actor performs value-guided policy improvement through the path-density objective, and the behavior prior supplies a recency-adaptive reference. As training proceeds, this reference changes with recent interaction data rather than remaining tied to a fixed dataset. The resulting objective assigns an additional cost to critic-favored actions with low density under the learned recent-behavior reference.

The resulting correction should be interpreted as an implementable reference-calibration term for online flow-policy learning. It is MI-motivated through the policy-reference log-ratio form, but it is not an exact mutual-information objective because the policy term is a flow-path density and the reference term is a learned recent-behavior prior rather than the exact current-policy marginal. This interpretation matches the goal of MIFLOW: calibrating critic-guided updates under finite and evolving online interaction data.

\section{Experiments}

We evaluate MIFLOW on MuJoCo continuous-control benchmarks to examine whether a recency-adaptive behavior prior can calibrate critic-guided updates of expressive flow policies under finite and evolving online interaction data.
Specifically, we ask three questions.
First, how does MIFLOW compare descriptively with the evaluated SAC-style and flow-policy baselines on six MuJoCo-v4 tasks?
Second, do prior-defined diagnostics change in a direction consistent with stronger policy alignment to the learned recent-behavior reference?
Third, how do the observed results vary across the evaluated prior families and prior-buffer sizes?

\subsection{Experimental Setup}

\noindent\textbf{Environments.}
We evaluate all methods on six MuJoCo continuous-control tasks: Ant-v4, HalfCheetah-v4, Hopper-v4, Humanoid-v4, HumanoidStandup-v4, and Walker2d-v4. These tasks cover different action dimensions, locomotion dynamics, and optimization difficulty, making them suitable for evaluating expressive online policies under continuous control.

\noindent\textbf{Baselines.}
We compare MIFLOW with three baselines in the main performance evaluation. SAC is included as a standard maximum-entropy actor-critic method with a Gaussian policy \cite{haarnoja2018soft,haarnoja2019softactorcriticalgorithmsapplications}. SACFlow is included as a path-density flow-policy actor-critic baseline without the proposed recency-adaptive reference \cite{zhang2025sac}. FlowRL is included as a recent online flow-policy method \cite{lv2025flowrl}.

For diagnostic and ablation studies, we additionally compare prior variants within the same path-density flow-policy optimization framework. SACFlow denotes the no-prior baseline. SACFlow-GP denotes a learned Gaussian-prior variant. MIFLOW denotes the full method with a learned flow-matching behavior prior. These variants provide descriptive comparisons of adding a behavior prior and of using a flow-matching rather than a learned Gaussian prior.

\noindent\textbf{Training protocol.}
Main MuJoCo learning curves are plotted over \(1{,}000{,}000\) environment steps using five random seeds. For Fig.~\ref{fig:main_results}, curves show smoothed mean episodic return, and shaded regions denote standard deviation across seeds. For the prior-family ablation in Table~\ref{tab:prior_family_ablation}, we report the final evaluated episodic return as mean \(\pm\) SEM over five random seeds. The prior-buffer sensitivity in Table~\ref{tab:prior_buffer_ablation} reports the final evaluated episodic return and prior-defined diagnostics as mean \(\pm\) SEM over five random seeds. Unless otherwise stated, MIFLOW uses \(K=4\) flow sampling steps and a default prior-buffer size of \(5\times 10^4\). Baseline hyperparameters follow their corresponding implementations, and MIFLOW-specific settings are fixed across tasks unless otherwise stated. All comparisons are descriptive, with the number of seeds stated for each experiment. We do not conduct formal hypothesis tests, so differences between methods should not be interpreted as statistically established.

\noindent\textbf{Metrics.}
The primary metric is episodic return. For prior-defined diagnostics, OOS Rate measures the fraction of policy-generated actions whose action-dimension-normalized \(\log\rho_\phi(a)\) falls below the 20th percentile of recent-buffer actions under the same learned prior. This is a prior-defined diagnostic rather than an oracle measure of true state-action support. HQ/LP Cond measures the fraction of top-20\% sampled policy actions under clipped double-Q that also fall into this low-prior region. In these diagnostics, one policy action is sampled per diagnostic state.

\subsection{Main Results}

\begin{figure*}[t]
\centering
\includegraphics[width=\textwidth]{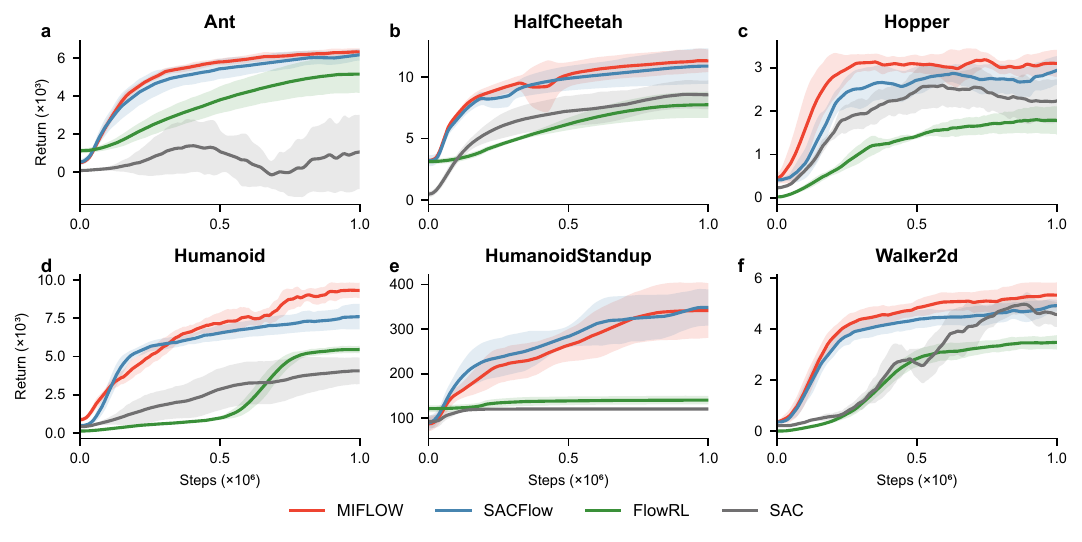}
\caption{
Main performance comparison on six MuJoCo-v4 continuous-control tasks.
Curves show smoothed mean episodic return over environment steps, and shaded regions denote standard deviation across five seeds.
These curves provide descriptive comparisons and do not indicate formal statistical significance.
}
\label{fig:main_results}
\end{figure*}

Figure~\ref{fig:main_results} presents descriptive learning-curve comparisons among MIFLOW, SACFlow, FlowRL, and SAC on six MuJoCo-v4 tasks. MIFLOW had higher late-training mean returns than the evaluated baselines on Hopper-v4, Humanoid-v4, and Walker2d-v4. On Ant-v4, its final mean return was close to that of SACFlow. MIFLOW and SACFlow produced similar mean curves on HalfCheetah-v4 and HumanoidStandup-v4, and both were generally above SAC and FlowRL during most of training. Because the comparisons use five seeds without formal hypothesis tests, these patterns should be interpreted as task-dependent descriptive results rather than statistically established differences.

Relative to SAC, MIFLOW retains a flow-based policy parameterization. Relative to SACFlow, it adds a recency-adaptive behavior prior as a dynamic reference. Relative to FlowRL, it uses a tractable path-density actor-critic formulation. The reported results characterize these methods within the evaluated MuJoCo protocol and do not establish broader superiority across online reinforcement-learning settings.

\subsection{Prior-Referenced Diagnostic Analysis}

\begin{figure}[t]
\centering
\includegraphics[width=\columnwidth]{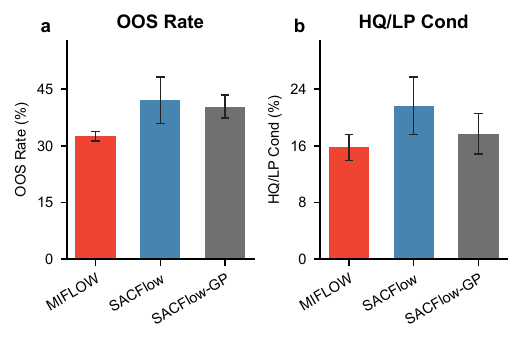}
\caption{
Walker2d-v4 diagnostic case study.
Bars report mean \(\pm\) SEM over five random seeds.
SACFlow denotes the no-prior path-density flow-policy baseline, and SACFlow-GP denotes the learned Gaussian-prior variant.
OOS Rate is prior-defined: it counts policy actions whose action-dimension-normalized \(\log\rho_\phi(a)\) falls below the 20th percentile of recent-buffer actions under the same learned prior.
HQ/LP Cond is the fraction of top-20\% clipped-double-Q policy actions that also lie in this low-prior region.
MIFLOW has the lowest reported mean for both prior-defined diagnostics in this case study.
}
\label{fig:diagnostic_results}
\end{figure}

We next examine how policy actions change relative to the learned recent-behavior reference. The two diagnostics use the same prior that enters the MIFLOW objective. They therefore measure policy-prior alignment, not true state-action support or critic extrapolation error. Figure~\ref{fig:diagnostic_results} presents a task-specific diagnostic comparison among the three variants on Walker2d-v4. Table~\ref{tab:prior_buffer_ablation} additionally reports the same diagnostics for the prior-buffer-size sensitivity study on Ant-v4 and Walker2d-v4. Together, these analyses provide internal consistency checks of the proposed reference mechanism.

On Walker2d-v4, MIFLOW had the lowest mean prior-defined OOS Rate and HQ/LP Cond among the three evaluated variants. These values indicate closer alignment with the learned recent-behavior reference and fewer high-Q samples in regions assigned low density by that reference. Because the diagnostics are defined by the learned prior, they cannot distinguish reduced critic extrapolation from improved consistency with the chosen prior model. The result is therefore descriptive and task-specific.

\subsection{Ablation Studies}

We conduct ablation studies to examine two factors that directly affect the proposed recency-adaptive reference.
The first study compares the learned flow-matching behavior prior with a learned Gaussian-prior variant, testing whether the prior family matters for representing the recent action distribution.
The second study varies the recent-action prior-buffer size, testing how the recency timescale of the behavior reference affects performance and prior-defined diagnostics.

\begin{table}[t]
\centering
\footnotesize
\setlength{\tabcolsep}{4pt}
\caption{
Prior-family ablation on MuJoCo tasks.
Values report the final evaluated episodic return, summarized descriptively as mean \(\pm\) SEM over five seeds.
SACFlow-GP denotes the learned Gaussian-prior variant.
}
\label{tab:prior_family_ablation}
\begin{tabular}{lcc}
\toprule
Environment & MIFLOW \((\times 10^3)\) & SACFlow-GP \((\times 10^3)\) \\
\midrule
Ant & 6.3 \(\pm\) 0.1 & -0.8 \(\pm\) 0.3 \\
HalfCheetah & 11.3 \(\pm\) 0.5 & 9.0 \(\pm\) 0.7 \\
Hopper & 3.1 \(\pm\) 0.2 & 3.0 \(\pm\) 0.1 \\
Humanoid & 9.3 \(\pm\) 0.3 & 4.2 \(\pm\) 0.3 \\
HumanoidStandup & 341.8 \(\pm\) 30.7 & 99.6 \(\pm\) 14.9 \\
Walker2d & 5.3 \(\pm\) 0.2 & 5.0 \(\pm\) 0.3 \\
\bottomrule
\end{tabular}
\end{table}

Table~\ref{tab:prior_family_ablation} reports final mean returns for the flow-matching and Gaussian-prior variants. The flow-matching variant had higher mean returns on all six tasks, although the differences on Hopper-v4 and Walker2d-v4 were small relative to the reported SEMs. Larger descriptive separations occurred on Ant-v4, HalfCheetah-v4, Humanoid-v4, and HumanoidStandup-v4. These results are consistent with prior-family choice affecting training behavior in most evaluated settings, but they do not establish that flow-matching priors generally outperform Gaussian priors.

\begin{table}[t]
\centering
\scriptsize
\setlength{\tabcolsep}{2.5pt}
\caption{
Prior-buffer-size sensitivity on Ant-v4 and Walker2d-v4.
This is a separate sensitivity suite in which only the prior-buffer size is varied.
Values report mean \(\pm\) SEM over five random seeds.
Return is the final evaluated episodic return.
}
\label{tab:prior_buffer_ablation}
\begin{tabular}{llccc}
\toprule
Env & Buffer & Return \((\times 10^3)\) & OOS (\%) & HQ/LP (\%) \\
\midrule
Ant & \(5\mathrm{k}\) & 6.55 \(\pm\) 0.03 & 30.2 \(\pm\) 1.0 & 15.9 \(\pm\) 2.5 \\
 & \(50\mathrm{k}\) & 6.3 \(\pm\) 0.1 & 32.2 \(\pm\) 1.9 & 15.4 \(\pm\) 0.4 \\
 & \(500\mathrm{k}\) & 6.36 \(\pm\) 0.30 & 33.7 \(\pm\) 1.0 & 19.7 \(\pm\) 2.3 \\
\midrule
Walker2d & \(5\mathrm{k}\) & 5.42 \(\pm\) 0.41 & 37.6 \(\pm\) 2.3 & 10.2 \(\pm\) 0.6 \\
 & \(50\mathrm{k}\) & 5.3 \(\pm\) 0.2 & 37.9 \(\pm\) 2.3 & 16.1 \(\pm\) 4.1 \\
 & \(500\mathrm{k}\) & 5.50 \(\pm\) 0.45 & 35.4 \(\pm\) 4.7 & 17.4 \(\pm\) 6.0 \\
\bottomrule
\end{tabular}
\end{table}

Table~\ref{tab:prior_buffer_ablation} reports descriptive results for three prior-buffer sizes on Ant-v4 and Walker2d-v4. The point estimates do not follow a common monotonic pattern across return, prior-defined OOS Rate, and HQ/LP Cond. However, several differences are small relative to the reported SEMs, so these results do not establish a reliable non-monotonic dependence on buffer size. They only indicate that the selected buffer timescale can alter the learned reference and its associated diagnostics. The buffer size conceptually mediates a possible plasticity-stability trade-off, but the present comparisons are insufficient to characterize that trade-off empirically.

Overall, the prior-defined diagnostics summarize two related aspects of policy-prior alignment. They do not provide independent measurements of true support or critic extrapolation error.

\section{Limitations and Scope}

MIFLOW is evaluated only on six MuJoCo-v4 continuous-control tasks, so the reported results do not establish performance across other environment families or application domains. The behavior prior is state-independent and represents a recent action marginal rather than full state-action support. This simplification may be insufficient when admissible actions depend strongly on state. The reference correction is motivated by mutual-information and reference-regularization principles, but it is not an exact mutual-information objective or a formal robustness guarantee. Moreover, the prior-defined OOS Rate and HQ/LP Cond use the same learned prior that enters the training objective. They measure alignment with this reference rather than true support or critic extrapolation error. Diagnostic evidence is limited to a three-variant case study on Walker2d-v4 and a prior-buffer-size sensitivity study on Ant-v4 and Walker2d-v4. The main performance comparisons, prior-family ablation, diagnostic case study, and prior-buffer sensitivity each use five seeds; no formal hypothesis tests are conducted. The experiments also do not provide comprehensive sensitivity analyses for $\lambda_\rho$, prior-update frequency, or prior architecture, and they do not quantify the additional computational overhead of prior training and density evaluation. These questions remain important directions for broader evaluation.

\section{Conclusion}

This paper introduced MIFLOW, an online flow-policy method that uses a recency-adaptive behavior prior to calibrate path-density actor-critic updates. The prior provides a soft reference for recent action density and assigns an additional cost to critic-favored actions with low prior density. Across six MuJoCo-v4 tasks, MIFLOW produced task-dependent competitive or higher mean returns relative to the evaluated baselines. The Walker2d-v4 diagnostics and prior ablations further characterize how policy behavior varies relative to the learned reference. Within the stated experimental and diagnostic boundaries, these results support dynamic recent-behavior priors as a practical calibration mechanism for online flow-policy learning. Broader claims about critic extrapolation, statistical superiority, and generalization beyond MuJoCo require additional evidence.

\ifanonymous
\else
\section*{Author Contributions}
Chenhao Wang led the conceptualization, methodology, software, experiments, analysis, visualization, and initial drafting. Haonan Ma contributed to methodology, software, validation, visualization, and manuscript revision. Wenjie Zhang and Yifan Wang contributed to validation, result interpretation, and manuscript revision. Mingbin Gao, Yuan-Ye Zhang, and Wenjing Hong supervised the project and contributed to methodology, result interpretation, and manuscript revision. All authors reviewed and approved the final manuscript.

\section*{Acknowledgment}
This work was supported by the National Natural Science Foundation of China (22325303) and the Fujian Provincial ``Young Eagle Program'' for Outstanding Young Talent. The authors also acknowledge support from Eva AI for Science Lab.
\fi

\IEEEtriggeratref{13} 
\bibliographystyle{IEEEtran}
\bibliography{IEEEabrv,references}

\end{document}